\documentclass[letterpaper, 10 pt, conference]{ieeeconf}  % Comment this line out if you need a4paper

\IEEEoverridecommandlockouts                              % This command is only needed if 
\usepackage{times}
\usepackage{epsfig}
\usepackage{graphicx}
\usepackage{amsmath}
\usepackage{amssymb}
\usepackage{multirow}
\usepackage{hyperref}
\usepackage{color}
\usepackage{caption}
\usepackage{subcaption}
\usepackage{makecell}
\usepackage{mathtools}

\DeclarePairedDelimiter\abs{\lvert}{\rvert}
\title{\LARGE \bf
Benchmarking Pedestrian Odometry: The Brown Pedestrian Odometry Dataset (BPOD)
}

\author{David Charatan$^{*}$, Hongyi Fan$^{*}$, and Benjamin Kimia$^{*}$% <-this % stops a space
\thanks{This work was supported in part by NIH grant 5R01EY029745-03 and NSF Award 1910530.}% <-this % stops a space
\thanks{$^{*}$School of Engineering, Brown University, Providence, RI, USA}%
}

\begin{document}

\maketitle
\thispagestyle{empty}
\pagestyle{empty}
% \runninghead{Charatan, Fan, Kimia}{BPOD}

% Any macro definitions you would like to include
% These are not defined in the style file, because they don't begin
% with \bmva, so they might conflict with the user's own macros.
% The \bmvaOneDot macro adds a full stop unless there is one in the
% text already.
% \def\eg{\emph{e.g}\bmvaOneDot}
% \def\Eg{\emph{E.g}\bmvaOneDot}
% \def\etal{\emph{et al}\bmvaOneDot}
\def\numLocations{12\xspace}

%-------------------------------------------------------------------------
% Document starts here

%\maketitle
\begin{abstract}
We present the Brown Pedestrian Odometry Dataset (BPOD) for benchmarking visual odometry algorithms in head-mounted pedestrian settings. This dataset was captured using synchronized global and rolling shutter stereo cameras in 12 diverse indoor and outdoor locations on Brown University's campus. Compared to existing datasets, BPOD contains more image blur and self-rotation, which are common in pedestrian odometry but rare elsewhere. Ground-truth trajectories are generated from stick-on markers placed along the pedestrian's path, and the pedestrian's position is documented using a third-person video. We evaluate the performance of representative direct, feature-based, and learning-based VO methods on BPOD. Our results show that significant development is needed to successfully capture pedestrian trajectories. The link to the dataset is here: \url{https://doi.org/10.26300/c1n7-7p93}
\end{abstract}
\section{Introduction}
Visual Odometry (VO) is the process of measuring ego-motion using image data. Specifically, VO uses visual data to recover a navigating agent’s path relative to its position at an earlier time. This is in contrast to odometry based on other sensory data such as wheel sensors, step counters, global positioning system (GPS), IMUs, sonar, infrared, radio frequency (RF) receivers, laser range finders (LIDAR), RGB-D cameras, and others~\cite{Nister:etal:CVPR:2014,Scaramuzza:Fraundorfer:RA:2011, Fraundorfer:Scaramuzza:RA:2012}. Visual odometry has become predominant given the versatility and relatively low cost of cameras~\cite{MurArtal:etal:TR:2015, Tardos:ORBSLAM2:TR:2017, Newcombe:etal:DTAM:CVPR:2011, Engel:etal:PAMI:2017}. Challenges of ambiguous scale, motion blur during rapid rotations, low or repeated texture, and large dynamic range have encouraged the fusion of visual odometry with another low-cost and versatile sensor, namely, inertial (IMU) odometry; this is referred to as visual-inertial odometry (VIO)~\cite{Tsotsos:etal:ICRA:2015, Scaramuzza:Zhang:arXiv:2019}. Applications of visual odometry are vast and span planetary exploration, unmanned aerial vehicles (drones or MAVs),~\cite{Blosch:etal:ICRA:2010,Cvisic:etal:JFR:2018}, autonomous driving~\cite{Lategahn:etal:ICRA:2011,Geiger:etal:CVPR:2012}, augmented reality applications~\cite{Chekhlov:etal:SMAR:2007}, mobile mapping~\cite{Karam:etal:RS:2019}, service robotics~\cite{Shi:etal:ICRA:2020}, simultaneous localization and mapping (SLAM)~\cite{Bailey:Durrant:RA:2006, Durrant:Bailey:RA:2006, Cadena:etal:TR:2016}, {\em etc}.

The key to advancing the state of the art in VO is the availability of challenging, high-quality, broadly represented, and task-driven benchmarks. A case in point is the rapid development resulting from the introduction of KITTI~\cite{Geiger:etal:CVPR:2012} and other datasets~\cite{Pandey:etal:IJRR:2011,Maddern:etal:IJRR:2017,Yu:etal:arXiv:2018} for the task of autonomous driving. It is important to emphasize, however, that the utility of a benchmark is necessarily limited to the task for which it was designed. For example, autonomous driving benchmarks contain paths that are planar and mostly straight, with a small number of turns that have limited accelerations and radii of curvature. By contrast, other agents such as drones and pedestrians exhibit rapid rotations, high acceleration, and more general paths of motion~\cite{Antonini:etal:ISER:2018}.

Benchmarking is an elusive task with numerous subtleties. VO techniques can generally be classified into three categories, namely, feature-based methods~\cite{ORB-SLAM}, direct methods~\cite{Newcombe:etal:DTAM:CVPR:2011,Engel:etal:PAMI:2017} and deep learning methods~\cite{wang2017deepvo, li2018undeepvo, yang2020d3vo, zhao2020towards}, and certain datasets can be more suitable for one or the other. Textured, feature-rich scenes favor feature-based methods while texture-less scenes with large homogeneous areas favor direct methods. Illumination variation ({\em e.g.}, the requirement to illuminate darker environments like those found underground or underwater) impacts the photometric invariance assumption of direct methods~\cite{Kasper:etal:IROS:2019}. Similarly, a dataset whose images have not been photometrically calibrated ({\em i.e.,} where exposure times, the camera response function, and lens vignetting have been measured) disfavors direct methods. These and other nuances have led to an abundance of benchmarks with varying targets~\cite{KITTI_Dataset, EuRoC_MAV_Dataset, zuniga2020vi, pfrommer2017penncosyvio,TUM_Mono_Dataset, Carlevaris:etal:IJRR:2016,Majdik:etal:IJRR:2017}. Benchmarks are inherently task-oriented and must be constructed carefully to satisfy the requirement of the application at hand.  

\begin{figure}[t]
    \centering
    \includegraphics[width=\linewidth]{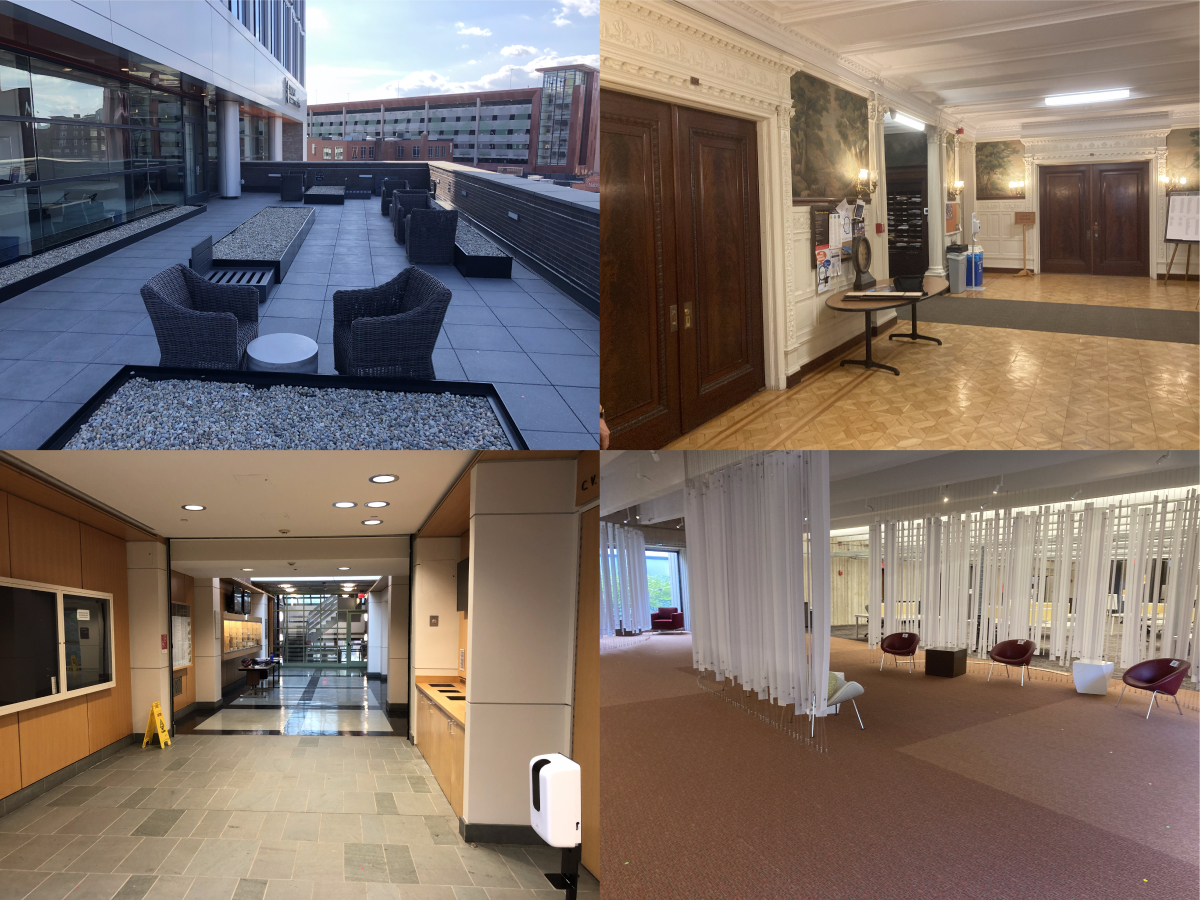}
    \caption{BPOD spans a diverse set of indoor and outdoor scenes ranging from texture-rich historic buildings to modern construction.}
    \label{fig:Teaser}
\end{figure}

\begin{table*}[ht]
    \centering
    \caption{Comparison of representative datasets.}
    \begin{tabular}{c|ccccc}
        Dataset & Mounting & Environment & Camera Type & Image & Ground Truth  \\
        \hline
        KITTI~\cite{KITTI_Dataset} & Car & Outdoors & Global Shutter & Stereo/Auto Exposure & GPS \\
        EuRoc~\cite{EuRoC_MAV_Dataset} & MAV & Indoors & Global Shutter & Stereo/Auto Exposure & Motion Capture \\
        TUM~\cite{TUM_Mono_Dataset} & Hand-held & Indoors & Global Shutter & Monocular/Auto Exposure & Motion Capture \\
        ICL-NIUM~\cite{ICL_NUIM_dataset} & Synthetic & Synthetic-Indoors & Synthetic & Monocular/Synthetic & Synthetic \\
        MineNav~\cite{MineNav_dataset} & Synthetic & Synthetic-Outdoors & Synthetic & Monocular/Synthetic & Synthetic \\
        UMA-VI~\cite{zuniga2020vi} & Hand-held & Indoor/Outdoors & Global Shutter & Stereo / Auto Exposure & SfM \\
        PennCOSY~\cite{pfrommer2017penncosyvio} & Hand-held & Indoor/Outdoors  & Mix & Stereo/Auto Exposure & Fiducial/SfM \\
        BPOD (Ours) & Head-mount & Indoor/Outdoors & Mix & Stereo/Mix Exposure & Marker
    \end{tabular}
    \label{tab:datasets}
\end{table*}

Pedestrian tracking from head-mounted cameras is an application for which VO datasets are not available to the best of our knowledge. Existing datasets are commonly vehicle-mounted~\cite{Geiger:etal:CVPR:2012,Blanco:etal:IJRR:2014,Jeong:etal:IJRR:2019}, Segway-mounted ~\cite{Carlevaris:etal:IJRR:2016}, MAV-mounted~\cite{Burri:etal:IJRR:2015,Majdik:etal:IJRR:2017}, or hand-held~\cite{Pfrommer:eta;ICRA:2017,Schubert:etal:IROS:2018,Cortes:etal:ECCV:2018,Kasper:etal:IROS:2019}. The rare exception is one sequence (Campus-run) obtained from a head-mounted Velodyne LIDAR, but not from visual cameras. The authors included this sequence to highlight the added importance of the IMU in head-mounted situations which depict erratic movements~\cite{Neuhaus:etal:GCPR:2018}. As such, the proposed dataset, the Brown Pedestrian Odometry Dataset (BPOD), fills a gap in a significant application area.

\begin{figure}
    \centering
    (a) \includegraphics[height=0.19\linewidth]{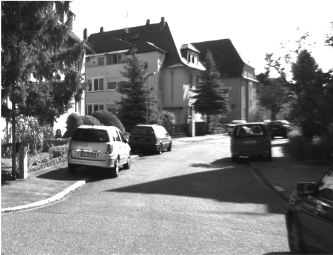}
    (b) \includegraphics[height=0.19\linewidth]{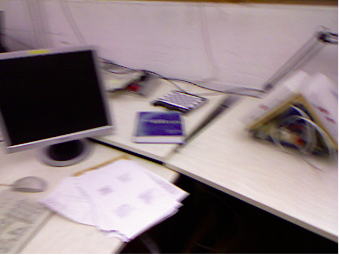}
    (c) \includegraphics[height=0.19\linewidth]{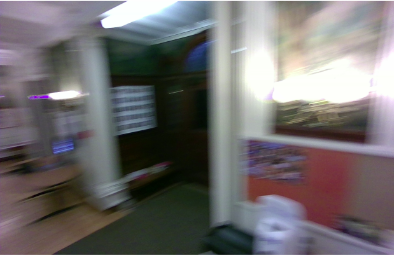}
    \caption{Typical turning sequences from vehicle-mounted (a), hand-held (b), and head-mounted (c) cameras demonstrate the potential for comparatively severe blur in head-mounted sequences.}
    \label{fig:intro}
\end{figure}

A head-mounted camera presents challenges not always present in the other datasets: inherent in a head-mounted camera are erratic and rapid head movements. Also, pedestrians often make quick rotations, which leads to blur in images, as shown in Figure~\ref{fig:intro}. 

The creation of a VO-focused dataset like BPOD requires several components: (i) A sensory platform for data acquisition: we selected two synchronized stereo cameras, one with a rolling shutter and one with a global shutter. Embedded within the latter is an IMU. In addition, one camera can acquire depth images, although the current version of BPOD omits these in favor of higher resolution images and higher frame rates. (ii) Video sequences: ours are obtained from indoor and outdoor scenes by a pedestrian wearing the helmet following a path annotated on the ground by small markers. An additional camera records the pedestrian's movement. (iii) Data processing to correlate pedestrian position to a map of markers to generate the ground-truth data. (iv) Experiments to evaluate the performance of VO algorithms: we tested three classes of odometry approaches on BPOD. The BPOD dataset is curated by the Brown University Library and is freely available for public use (the link will be included in the camera-ready version). Following a review of related datasets, this paper is organized according to these four components.

\section{Related Odometry Datasets}
Datasets focused on a variety of settings have been proposed to evaluate the performance of visual odometry systems. Perhaps the most well known is KITTI~\cite{KITTI_Dataset}, which focuses on outdoor car-mounted stereo imaging, with ground-truth trajectories obtained via GPS. While this dataset showcases a diverse set of driving scenes, it is inherently limited by its outdoor, vehicle-focused setting. The EuRoC MAV datset~\cite{EuRoC_MAV_Dataset} provides stereo images from a drone in an indoor environment. Its ground truth is captured by an external motion capture system and laser tracker. While this provides high-quality ground-truth trajectories, the necessary extra equipment requires a controlled environment, which prevents the creation of a large-scale dataset for indoor environments. The TUM RGB-D dataset~\cite{TUM_RGBD} was one of the first datasets focused on hand-held cameras. It features a large set of hand-held video sequences, but the ground-truth timestamps are not well aligned with the camera. The ground truth of this dataset is also captured by an external motion capture system, so this dataset only covers limited range of scenes. Additionally, the motion of the hand-held camera is slow and smooth, which is appropriate for hand-held scanning applications, but not for pedestrian ego-motion estimation. To overcome the limitations of external motion capture systems, ground truths based on fiducial markers have been used in a number of datasets~\cite{pfrommer2017penncosyvio}. However, these prominent markers generate extra feature correspondences within the scene (note that the markers used for our ground truths are non-invasive). Following the development of structure from motion (SfM) techniques, several datasets have used SfM to generate ground-truth camera trajectories. For example, \cite{zuniga2020vi}~and~\cite{majdik2017zurich} use COLMAP and Pix4D respectively to generate their ground truths. However, unlike these carefully constructed hand-held and MAV datasets, our dataset contains rapid blur and feature-less segments that make SfM-based ground truth generation infeasible. Other datasets omit ground-truth trajectories entirely, instead relying on loop closure. For example, the TUM MonoVO~\cite{TUM_Mono_Dataset} dataset, which features a large set of hand-held, photometrically calibrated, monocular footage, proposes an evaluation metric based on loop closure to evaluate VO without ground-truth trajectories.

\begin{table*}[t]
\centering
\caption{The streams we recorded for each sensor.}
\begin{tabular}{c|c|c|c|c|c|c|c|c}
\hline
Camera & Sensors & Resolution  & Frame Rate (Hz) & Sensor Aspect Ratio & Focal Length & FOV & baseline & Shutter Type\\ \hline
\multirow{2}{*}{D455} & Color             & 1280 \texttimes 800  & 30                                    & 16:10 & 1.88mm & 77$^{\circ}$ & 95mm & \multirow{2}{*}{Global} \\ \cline{2-7}
                      & Stereo Monochrome & 1280 \texttimes 800  & 30                                   &  8:5 & 1.93mm & 100.6$^{\circ}$ &   &              \\ \cline{1-9}
\multirow{2}{*}{D415} & Color             & 1920 \texttimes 1080 & 30                                    & 16:9 & 1.88mm & 77$^{\circ}$ & 55mm & \multirow{2}{*}{Rolling} \\ \cline{2-7}
                      & Stereo Monochrome & 1280 \texttimes 720  & 30     &                       16:9    & 1.88mm & 77$^{\circ}$  &  &  \\ \hline
\end{tabular}
\label{table:streams}
\end{table*}

In the face of these difficulties in generating ground truths for real cameras, synthetic datasets have also been devised for VO development. The ICL-NUIM dataset~\cite{ICL_NUIM_dataset} consists of 8 sequences of photorealistic synthetic indoor scenes in a monocular setting. The MineNav dataset~\cite{MineNav_dataset} presents an outdoor odometry dataset based on the sandbox game Minecraft. However, it is hard for these synthetic datasets to simulate features of real scenes, such as motion blur and pedestrian head bobbing. A summary of representative datasets is listed in Table~\ref{tab:datasets}.

\section{BPOD Sensory Platform}
\label{sec:data-acquisition}

\begin{figure}[h]
\centering
\begin{subfigure}[t]{\linewidth}
\includegraphics[width=\linewidth]{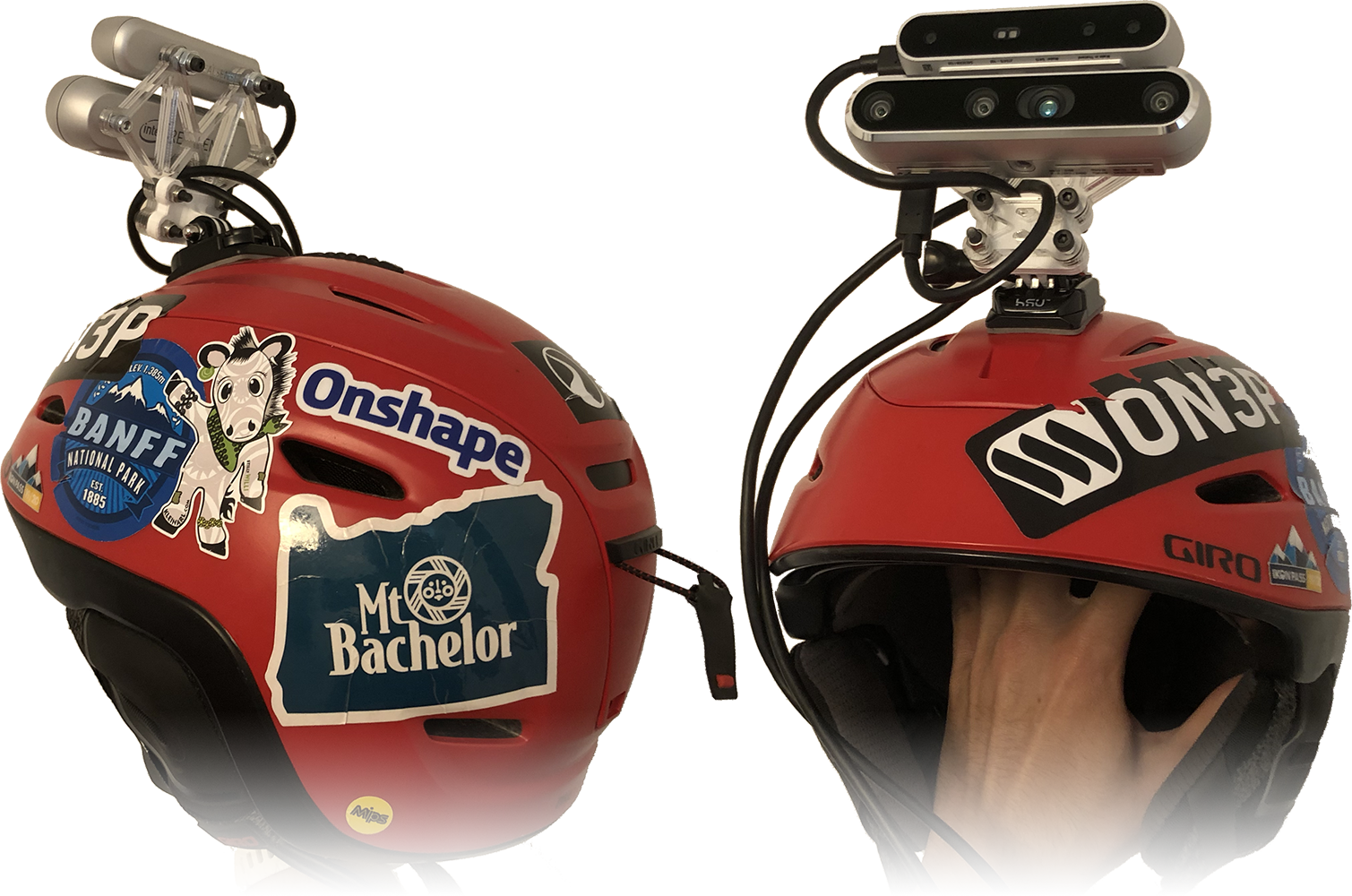}
\end{subfigure}

\begin{subfigure}[t]{\linewidth}
\includegraphics[width=\linewidth]{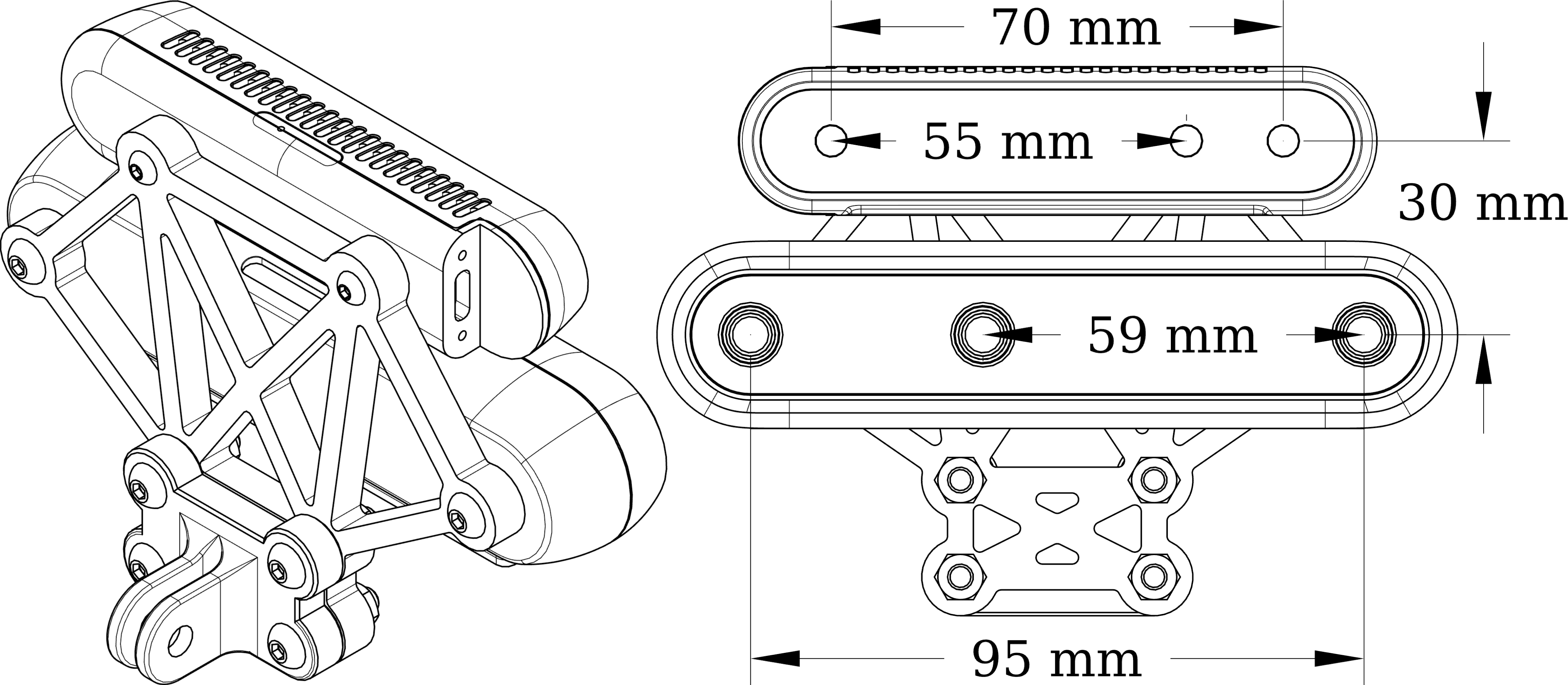}
\end{subfigure}

\caption{Two views of the camera assembly, shown together with frontal inter-camera measurements and CAD drawings.}
\label{fig:helmet}
\end{figure}

The camera mount most appropriate for pedestrian visual odometry is a head mount: cameras with chest-mounted or hand-held configurations are often occluded by the subject's arms or the presence of other pedestrians. The BPOD camera rig is a ski helmet outfitted with a standard GoPro mount. Figure~\ref{fig:helmet} shows the 3D-printed frame we used to attach our cameras to the helmet.

Our camera selection was guided by several constraints: We required {\em (i)} synchronized stereo cameras to compare monocular and stereo configurations for pedestrian odometry; {\em (ii)} high image quality, resolution, and frame rate; {\em (iii)} exposure control to allow for precise photometric calibration; {\em (iv)} both rolling shutter and global shutter cameras for comparison; {\em (v)} a universal driver for compatibility with a variety of devices; {\em (vi)} an IMU; and {\em (vii)} consumer-level pricing. 

Given these constraints, we chose a pair of Intel RealSense cameras~\cite{keselman2017intel} whose specifications are outlined in Table~\ref{table:streams}. An additional advantage of this choice is the ability to capture depth maps, although we could not incorporate this feature into BPOD due to data transfer bandwidth constraints. A few competing choices were {\em (i)} oCam from WithRobot~\cite{oCam}, which did not have exposure control and had inferior image quality; {\em (ii)} an ArduCam~\cite{ArduCam} which was limited to Nvidia Jetson or Raspberry Pi interfaces; and {\em (iii)} BumbleBee2~\cite{bee}, a global shutter stereo camera which was discontinued by the manufacturer. 

\begin{figure}[h]
\centering
\includegraphics[width=\linewidth]{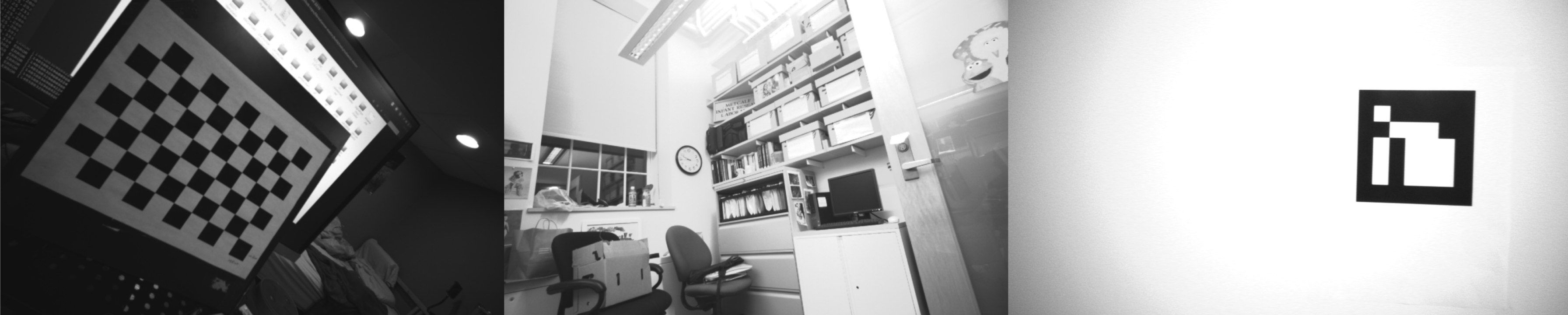}
\caption{Left to right: A sample checker board image, an image in the sweep exposure set, and a sample vignette image used for calibration.}
\label{fig:calibration}
\end{figure}

\noindent \textbf{Intrinsic and Photometric Calibration:} We calibrated the intrinsic parameters, distortion parameters, photometric parameters and vignette parameters for each camera independently. For each camera, we captured three calibration sequences: {\em (i)}~A $10 \times 7$ checker board for intrinsic calibration; {\em (ii)} a static scene with sweeping exposure for photometric calibration, and {\em (iii)} An image sequence with an ArUco tag~\cite{garrido2014automatic} for vignette calibration, as shown in Figure~\ref{fig:calibration}. We estimated the intrinsic parameters and distortion parameters using the MATLAB calibration toolbox. We estimated the photometric and vignette parameters using the code provided by~\cite{TUM_Mono_Dataset}. The extrinsic calibration for stereo cameras consists of the baseline listed in Table~\ref{table:streams}. We extracted the relative pose between D455 and D415 camera from the design of the camera rig shown in Figure~\ref{fig:helmet}.

\noindent \textbf{IMU Calibration: } We used the RealSense Toolkit for IMU calibration to obtain intrinsic and extrinsic IMU calibrations, which are available in BPOD.
% of cameras, since there is only translation between cameras, so we directly use the baseline listed in the datasheet of RealSense camera. The baseline is already listed in Table~\ref{table:streams}. The translation between D455 and D415 camera is extracted from the design of the camera rig, listed in Figure~\ref{fig:helmet}
\begin{table}[h]
    \centering
     \caption{Basic Information of BPOD dataset}
     \hskip-0.1cm
    \begin{tabular}{c|c|c|c|c}
    Location (Abbrev.)     & \multicolumn{4}{c}{Length of Sequences (sec.)} \\
    \hline
    Applied Math Building (APMA)  &  86 &  86 & 84 & 84\\
    85 Waterman St. (BERT)    & 102 & 96 & 106 & 100 \\
    Brown Design Workshop (BDW) & 109 & 113 & 110 & 109  \\
    CIC Office Balcony (CIC) &  122 & 141 & 140 & 145 \\
    Center of Info. Tech. (CIT 2nd Floor)  & 75 & 73 & 80 & 80 \\
    Center of Info. Tech. (CIT 4th Floor)  & 91 & 91 & 96 & 94 \\
    Lobby of Eng. Res. Center (ERC) &   109 & 110 & x & x  \\
    Lobby of Friedman Hall (Friedman) & 95 & 97 & 99 & 98  \\
    Lobby of MacMillan Hall (MacMillian) & 95 & 90 & 96 & 93\\
    Science Lib. (SciLiTables) & 68 & 67 & 69 & 74 \\
    Science Lib. (SciLiShutters) & 84 & 82 & 100 & 84
    \end{tabular}
    \label{tab:locations}
\end{table}
\section{Data Acquisition Protocol}
\label{subsec:mapping-protocol}
BPOD sequences were recorded at 12 locations on Brown University's campus, as outlined in Table~\ref{tab:locations}. We selected these locations to encompass a wide range of architectural styles, lighting conditions, and texture properties.

\begin{figure}[t]
\centering
\begin{subfigure}[t]{\linewidth}
\includegraphics[width=\linewidth]{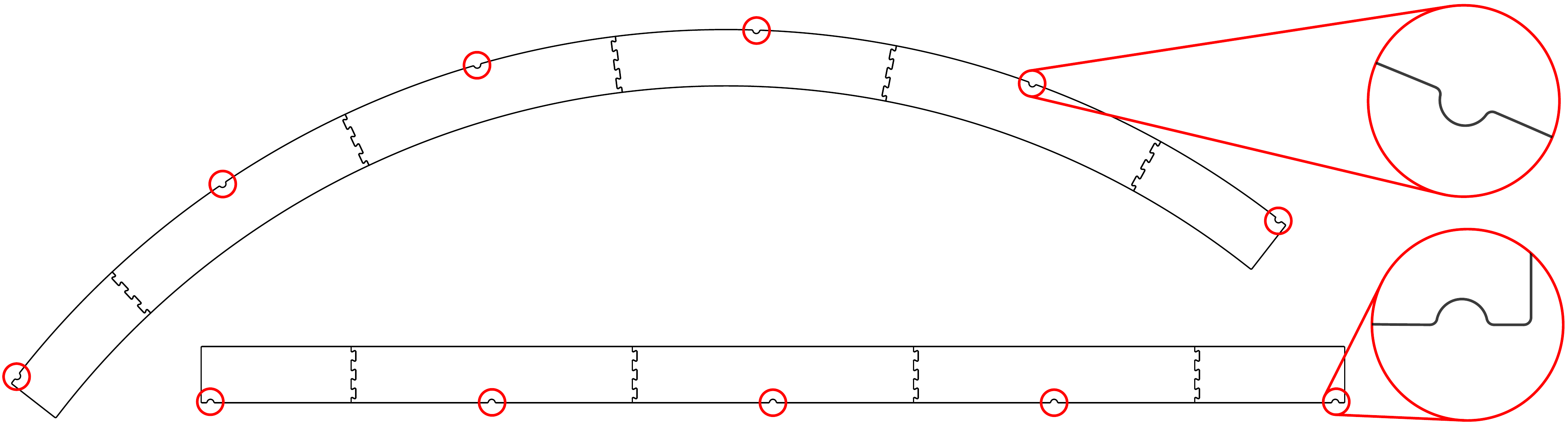}
\caption{CAD drawings of the templates used to place markers. Indentations for marker placement, which are spaced 30 inches apart, are highlighted and shown in detail views.}
\end{subfigure}

\begin{subfigure}[t]{\linewidth}
\includegraphics[width=\linewidth]{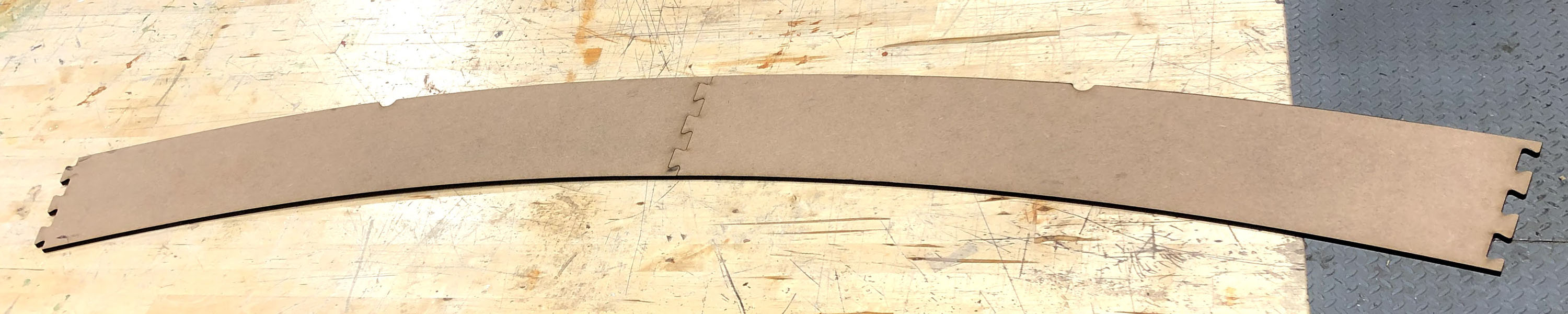}
\caption{Two laser-cut components for an arc template.}
\end{subfigure}

\begin{subfigure}[t]{.48\linewidth}
\includegraphics[width=\linewidth]{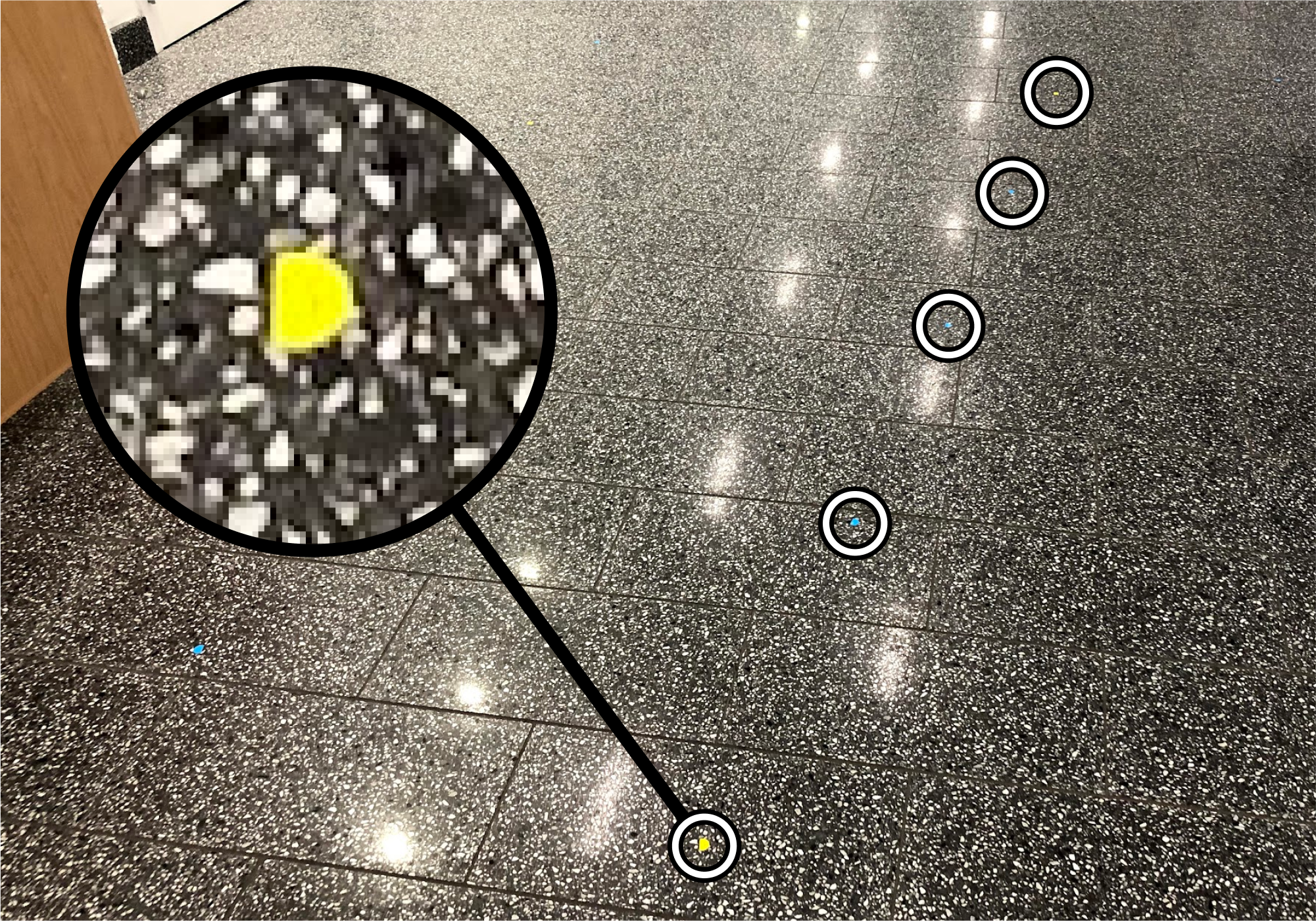}
\caption{Markers placed using the straight path segment. A key marker is highlighted.}
\end{subfigure}
\hfill
\begin{subfigure}[t]{.48\linewidth}
\includegraphics[width=\linewidth]{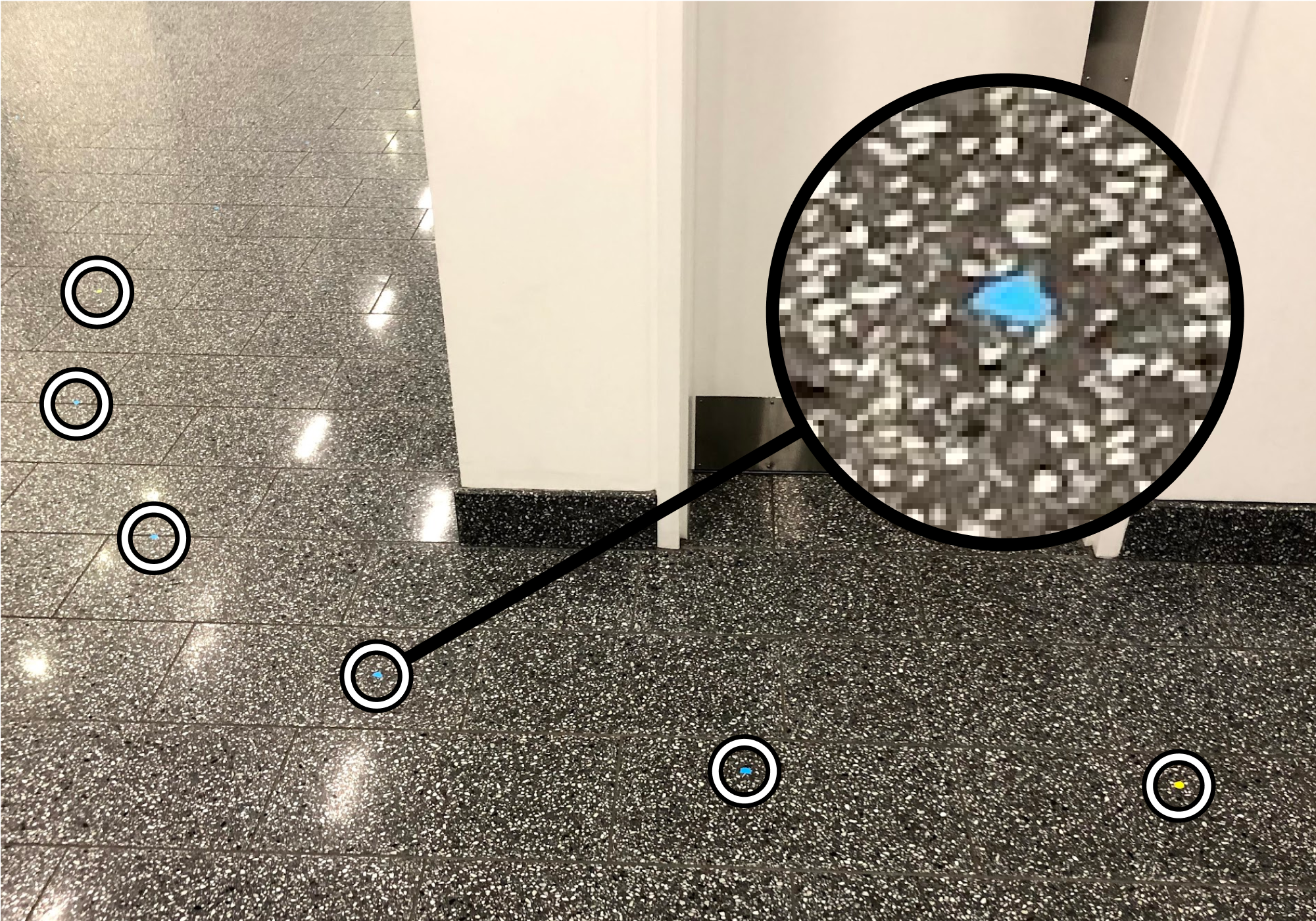}
\caption{Markers placed using the arc segment. An intermediate marker is highlighted.}
\end{subfigure}

\caption{Laser-cut templates and stick-on markers are used to generate ground-truth trajectories.}
\label{fig:templates}
\end{figure}

\noindent {\bf Defining Trajectories:} We used sequences of sticker markers to define intended trajectories for our video sequences. We placed these markers at 30-inch intervals using two laser-cut templates as shown in Figure~\ref{fig:templates}. One was a 120-inch straight template with slots for four markers, and a 90-degree circular arc template with slots for five markers. Our trajectories are closed loops, which allows the subject to traverse them multiple times in succession.

\noindent {\bf Mapping Trajectories:} We used pairwise distance measurements between markers in conjunction with triangulation-based initialization and descent-based optimization to construct two-dimensional trajectory maps. We partitioned the markers into key markers and intermediate markers. Key markers used a distinct sticker color and were placed at the ends of templates, while intermediate markers filled the templates' middle slots. From each key marker, we recorded distance measurements to all other non-occluded key markers using a hand-held Bosch Blaze Pro GLM165-40 laser distance measurer. Typically, about half of the other key markers were visible.

To convert sets of pairwise distance measurements to usable trajectories, we first generated coarse maps using classical triangulation. Recall that given the points $p_1$ at $(0, 0)$ and $p_2$ at $(u, 0)$, the position $(x, y)$ of a third point $p_\text{unknown}$ with distances $r_1$ to $p_1$ and $r_2$ to $p_2$ is given by the following equations:

An additional distance measurement between $p_\text{unknown}$ and a known position resolves $y$'s ambiguous sign. To apply this principle, we construct a graph $G$ where each key marker is a node, and each pairwise measurement between key markers forms an edge. Next, we form a subgraph $G_\text{mapped}$ from an arbitrary clique of size 3. We assign the first two markers positions of $(0, 0)$ and $(u, 0)$ respectively, where $u$ is the measured distance between the markers. Next, we use Equation~\ref{eq:triangulation} to determine the third marker's position, assuming a positive value of $y$. We then iteratively add each remaining node in $G$ to $G_\text{mapped}$ when at least three of its neighbors are in $G_\text{mapped}$, triangulating nodes' positions as they are added. If all markers can be triangulated, the initialization is complete; otherwise, we retry using a different initial clique. Note that because we assume a positive value of $y$ when triangulating the initial clique, we must manually reflect some of the resulting initializations about an arbitrary axis.

\begin{figure}[h]
    \centering
    \includegraphics[width=\linewidth]{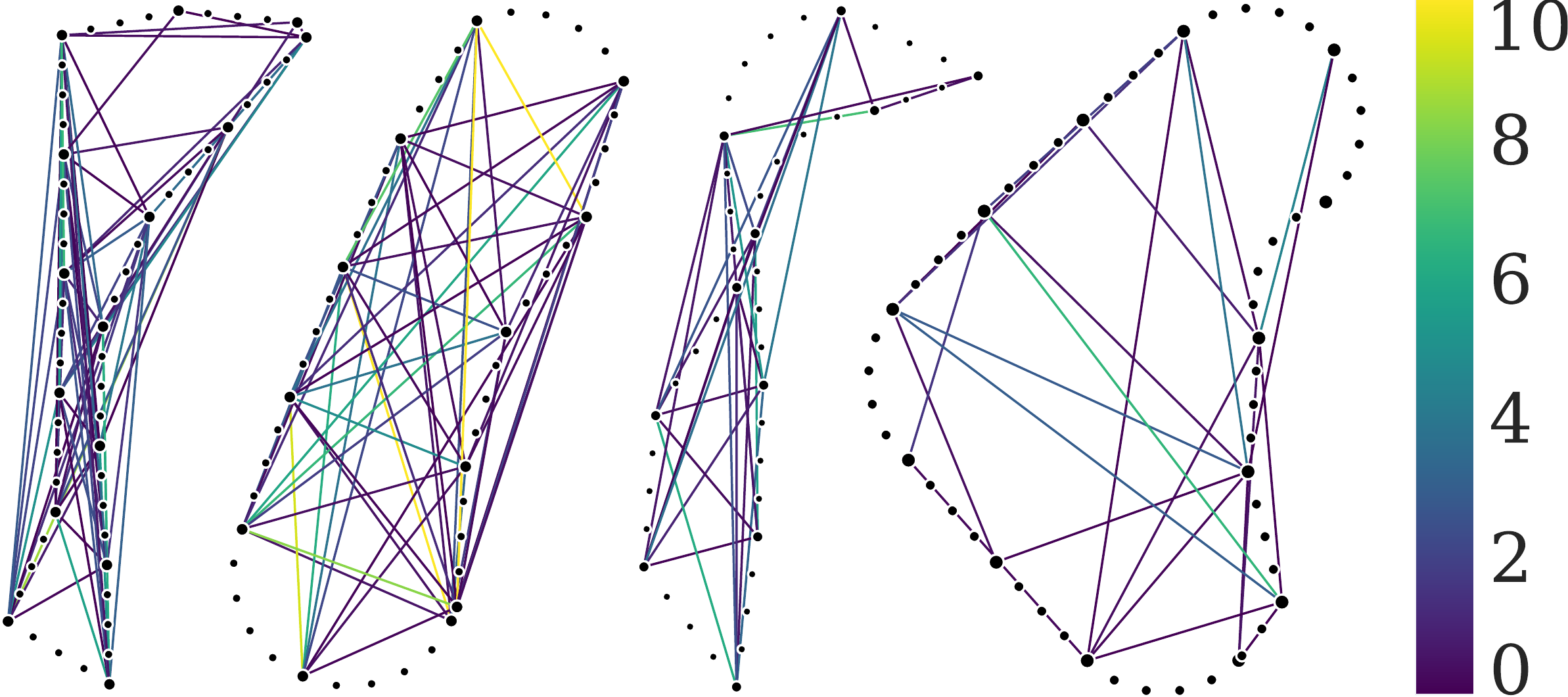}
    \caption{Maps showing measurements with their post-optimization errors for four locations. The legend's unit is millimeters.}
    \label{fig:measurement}
\end{figure}

Once initialization is complete, we refine the map using gradient descent. We optimize the markers' positions using an L1 loss $ L = \sum_i {\abs{d_i - \hat{d}_i}}$ between the measured distance $d_i$ and implied pairwise distances $\hat{d}_i$ between markers. This optimization generally leads to errors under 1cm (see Figure~\ref{fig:measurement}).

\noindent {\bf Discussion:} Compared to more complex and costly ground truth generation schemes, {\em e.g.}, laser trackers~\cite{delmerico2019we}, our scheme trades temporal resolution and the availability of 6-DOF information for simplicity. However, we argue that a ground truth's most important characteristic is that it provides accurate positional information over long distances, as this is essential for evaluating drift in odometry.
\begin{figure}[h]
    \centering
    \includegraphics[width=0.48\linewidth]{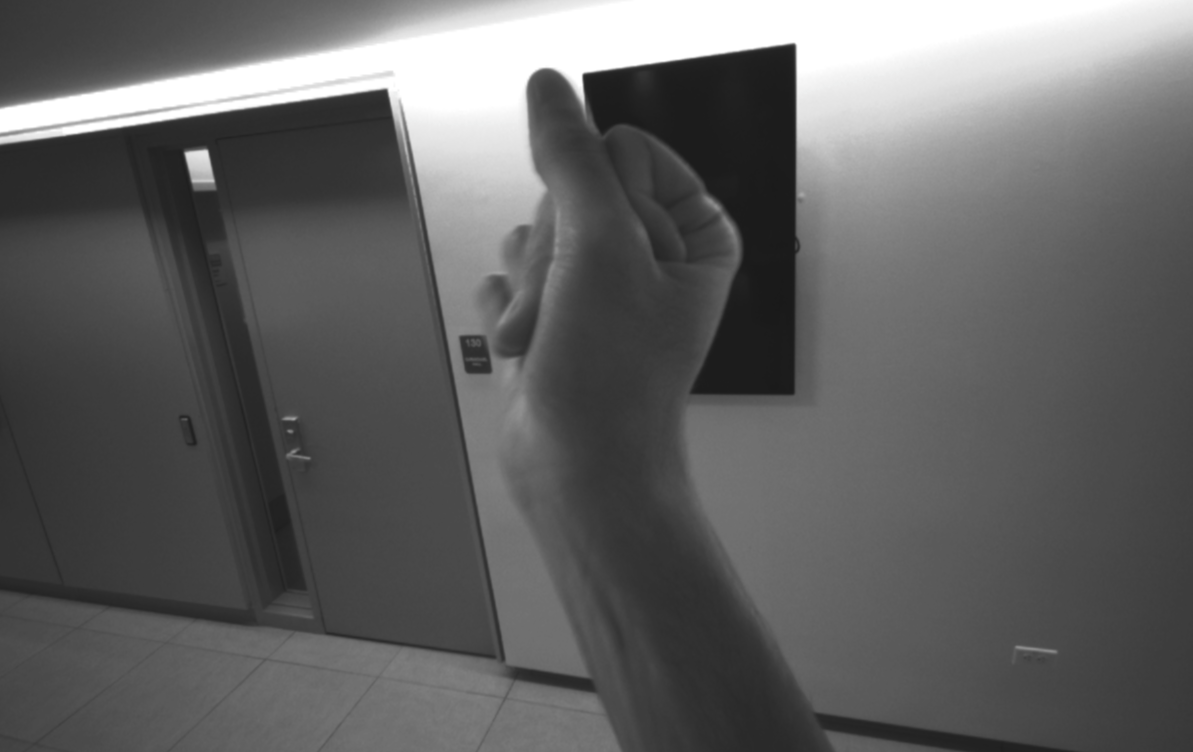}
    \hfill
    \includegraphics[width=0.48\linewidth]{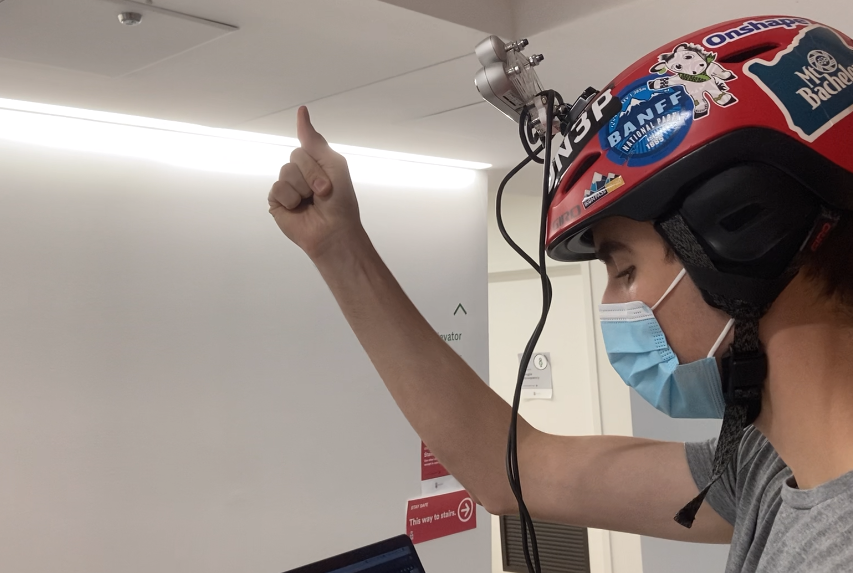}
    \caption{Finger snap synchronization of first-person (left) and third-person (right) video streams.}
    \label{fig:snap}
\end{figure}
\begin{figure*}[h]
    \centering
    \includegraphics[height=0.23\linewidth]{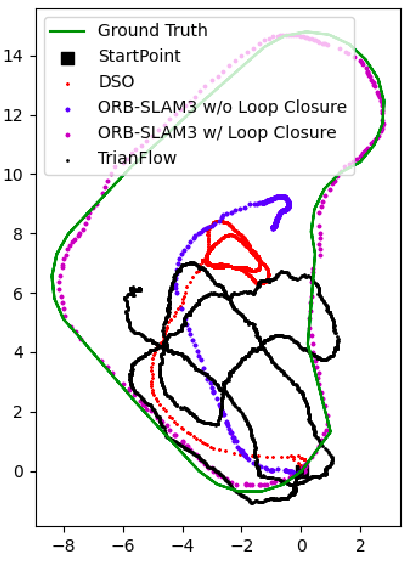}
    \includegraphics[height=0.23\linewidth]{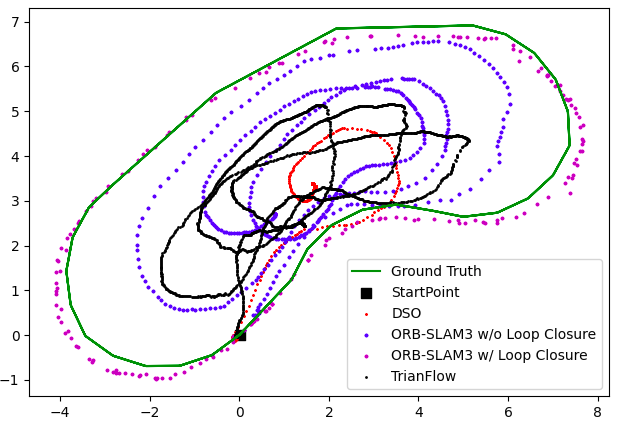}
    \includegraphics[height=0.23\linewidth]{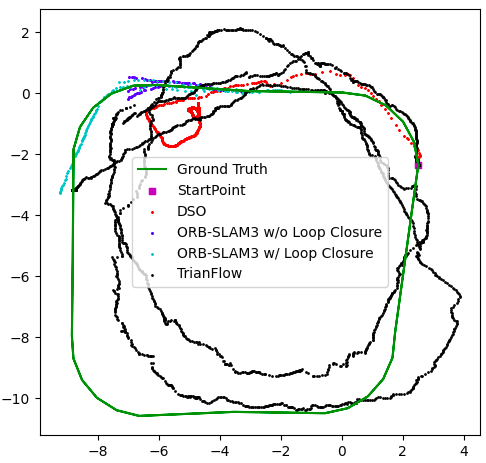}
    \includegraphics[height=0.23\linewidth]{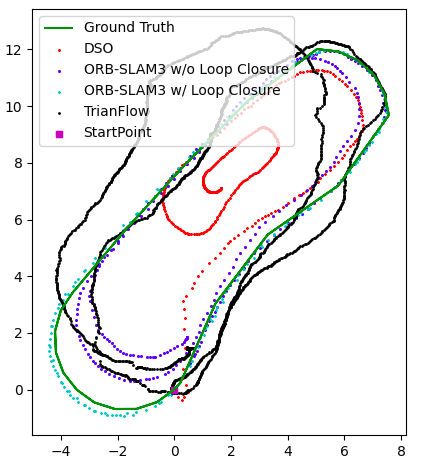}
    \caption{The estimated camera trajectories for four sample sequences.}
    \label{fig:trajecs}
\end{figure*}
\noindent {\bf Capturing video sequences:} We used a Macbook Pro to record data from the camera rig. We additionally recorded third-person video of the subject's feet with a smartphone to help reconstruct ground-truth trajectories. We synchronized the RealSense cameras and smartphone by starting each sequence with a finger snap, which is fast enough to ensure synchronization to within one frame, Figure~\ref{fig:snap}. 

\begin{equation}
\label{eq:triangulation}
\begin{aligned}
    x = \frac{r_1^2 - r_2^2 + u^2}{2u} \qquad y = \pm \sqrt{\abs{r_1^2 - x^2}}
\end{aligned}
\end{equation}

We captured four videos for each trajectory along two axes of variation: forward vs. backward trajectory traversal and fixed vs. auto exposure. Each trajectory follows a loop that is traversed 2 to 4 times. 

The idea of varying exposure is to probe the performance of various methods on scenes with a very high dynamic range of illumination. For example, the sequence ``4th floor of CIT'' transitions between a brightly lit atrium with skylights and a dimly lit set of hallways. We noted that the default setting of the auto-exposure mode is set relatively low, perhaps because it is intended to work with a laser projector. For the manually set exposure, we chose settings to balances overexposed (blown-out) regions and underexposed (completely black) regions.

\noindent {\bf Extracting the data:} Each sequence produces two distinct Robot Operating System (ROS) bag files, one for D415 and one for D455, which contain all data. We used a Docker container running an Ubuntu image with ROS to extract PNG images, camera calibration parameters, and IMU data along with the corresponding timestamps.

\noindent {\bf Image Position Annotation:} Our third-person smartphone videos allow us to identify the subject's location with respect to the marker-defined map at each point in time. We annotated the times at which the subject crosses individual markers, then interpolated the corresponding times and positions to produce fine-grained ground truth data. Our ground-truth data defines time relative to the first frame after the snap. This allows the two cameras' video streams to easily be synchronized with the smartphone's video.

\section{Experiments}
\label{sec:evaluation}
In this section, we describe the performance of three categories of visual odometry algorithms on BPOD. We selected a representative algorithm for each category: ORB-SLAM3~\cite{ORB-SLAM3} for feature-based approaches, DSO~\cite{DSO} for direct approaches, and TrianFlow~\cite{zhao2020towards} (trained on TUM dataset) for deep learning approaches. Both TrianFlow and DSO are restricted to monocular inputs, so we ran all tests using ORB-SLAM3's monocular mode. We use the authors' implementations of their odometry algorithms for all experiments. 

The evaluation of monocular odometry must address the inherent scale ambiguity in the resulting reconstructions: metric ambiguity in reconstruction and pose implies that scale as well as translation and rotation need to be matched optimally to compare two trajectories, {\em i.e.} two trajectories must by aligned under a similarity transformation. Observe that the ground truth of BPOD is in 2D while the computed odometry paths are in 3D. The latter are in fact fairly planar however, with small pseudo-sinusoidal depth variations that correspond to the bobbing head movements inherent to a pedestrian gait. Figure~\ref{fig:trajecs} compares the ground-truth trajectory to the reconstructed paths for each of the three approaches under the optimal similarity alignment constrained to matched start points ({\em i.e.} minimum distance under all relative relative rotations and scaling). It is clear that the reconstructed trajectories are not suitable for use in applications. In fact, these results are selected from among the very few sequences for which all methods completed the reconstruction of the trajectory!

A quantitative characterization of these results is shown in Table~\ref{tab:my_label}. A global evaluation metric $d_e$ measures the distance between the ground-truth and the estimated trajectories' end points. Observe that only three of twelve locations are completed by ORB-SLAM, in contrast to DSO which completes over 85\% of the locations and to TrianFlow which completes all; the endpoint error, $d_e$, however, is quite high for DSO and TrianFlow. 

\begin{figure}[h]
    \centering
    \includegraphics[width=\linewidth]{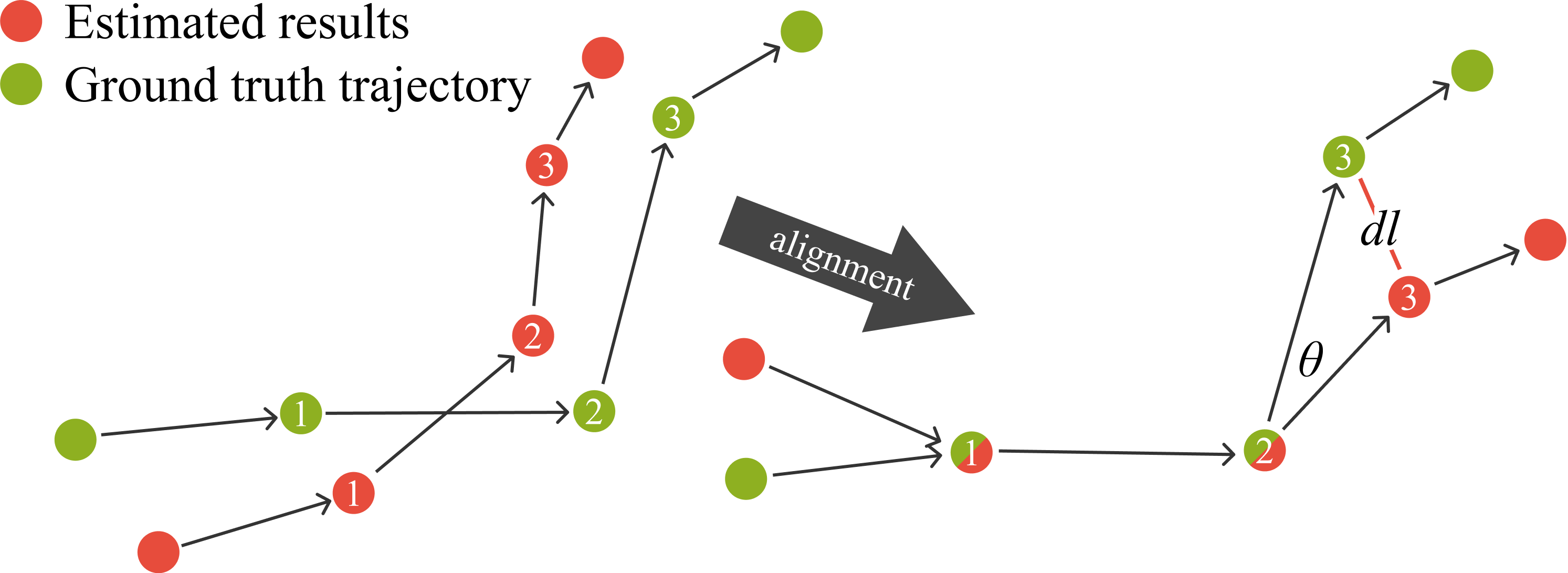}
    \caption{An illustration of our local error metrics.}
    \label{fig:metric}
\end{figure}

\begin{table*}[h]
    \centering
    \caption{Benchmarks for Each Location (\textcolor{red}{X}: Lost Track; -: Estimated Trajectory is short)}
    \begin{tabular}{|c|c|c|c|c|c|c|c|c|c|c|c|c|}
\hline
\multirow{2}{*}{Locations} & \multicolumn{3}{c|}{ORB-SLAM w/o LC} & %
    \multicolumn{3}{c|}{ORB-SLAM w/ LC} & \multicolumn{3}{c|}{DSO} & \multicolumn{3}{c|}{TrianFlow}\\
\cline{2-13}
 & $d_{e}$(m)&  $\theta$($^{\circ}/s$)& $d_{l}$($m/s$) & $d_{e}$(m)& $\theta$($^{\circ}/s$) & $d_{l}$($m/s$) & $d_{e}$(m)&  $\theta$($^{\circ}/s$) &  $d_{l}$($m/s$)& $d_{e}$(m) & $\theta$($^{\circ}/s$)& $d_{l}$($m/s$)\\
\hline
 APMA & \textcolor{red}{X} & 1.44 & 0.09 & \textcolor{red}{X} & 1.15 & 0.09 & 3.73& 4.13 & 0.27 & 4.20 & 8.80 & 0.43\\
\hline
 BDW & \textcolor{red}{X} & - & - & \textcolor{red}{X} & - & - &  7.62 & 3.23 & 0.22 & 6.48 & 6.27 & 0.42\\
\hline
BERT & \textcolor{red}{X} & 0.83 & 0.12 & \textcolor{red}{X} & 0.52 & 0.12 & 4.25 & 14.2 & 0.69 & 7.29 & 7.75 & 0.53\\
\hline
CIC Balcony & \textcolor{red}{X} & 6.3 & 0.63 & \textcolor{red}{X} & 7.00 & 1.99 & 4.33& 8.64 & 0.11 & 5.21 & 8.22 &0.67\\
\hline
CIT 2nd Floor & \textcolor{red}{X} & - & - & \textcolor{red}{X} & - & - & 7.02 & 3.88 & 0.68 & 8.86& 5.87 & 0.40 \\
\hline
CIT 4th Floor & 6.50 & 4.15 & 0.18 & 0.05 & 1.07 & 0.22 & 7.79 & 1.78 & 0.21 & 7.82 & 3.75 & 0.30 \\
\hline
ERC & \textcolor{red}{X} & 6.02 & 0.18 &  \textcolor{red}{X} & 5.96 & 0.18 & 6.95 & 3.00 & 0.48 & 8.73& 5.62 & 0.41\\
\hline
Friedman Hall & \textcolor{red}{X} & 4.55 & 0.23 & \textcolor{red}{X} & 5.94 & 0.23 & \textcolor{red}{X} & 4.53 & 0.45 & 9.92 & 10.77 & 0.54 \\
\hline
MacMillan & \textcolor{red}{X} & - & - & \textcolor{red}{X} & - & - & \textcolor{red}{X} & 5.57 & 0.57 & 9.87 & 8.18 & 0.57\\
\hline
SciLiShutters & 3.74 & 0.72 & 0.11 & 0.15 & 0.54 & 0.06 & 5.90 & 3.57& 0.50& 4.22& 5.10 & 0.56 \\
\hline
SciLiTables & 1.39 & 0.40 & 0.06 & 0.09 & 0.41 & 0.06 & 5.34 & 0.80 & 0.14 & 5.45 & 2.73 & 0.41 \\
\hline
MEAN & 3.88& 3.05 & 0.2 & 0.10 & 2.82 & 0.37 & 5.88 & 4.85 & 0.39 & 7.09 & 6.64 & 0.48\\
\hline
\end{tabular}
    \label{tab:my_label}
\end{table*}

A closer look at the reconstructed path reveals that scale drift is a significant issue, especially for DSO. This implies that a single scaling of the reconstruction path will lead to a large $d_e$ even though the shape is correctly estimated. We propose to use a local metric to decouple the error caused by scale drift from the error in reconstructing the shape of the path. Figure~\ref{fig:metric} illustrates the idea: consider the samples along each path, say one second apart. Aligning the first two samples adjusts the local scale of the reconstructed path to be that of the ground truth. Then, difference between the third point, angular difference $\theta$ and magnitude difference $d_l$, reveral the local reconstruction error. The conclusion is {\em (i)} ORB-SLAM is the most accurate in reconstruct a trajectory, but it frequently fails to do so; this is due to fast rotations that cause motion blur and the subsequent loss of feature tracking. {\em (ii)} DSO frequently completes the trajectory but suffers from scale drift despite accurate calibration and use of global shutter cameras; {\em (iii)} TrianFlow completes all trajectories but is the least accurate. 

\begin{figure}
    \centering
    \includegraphics[width=0.48\linewidth]{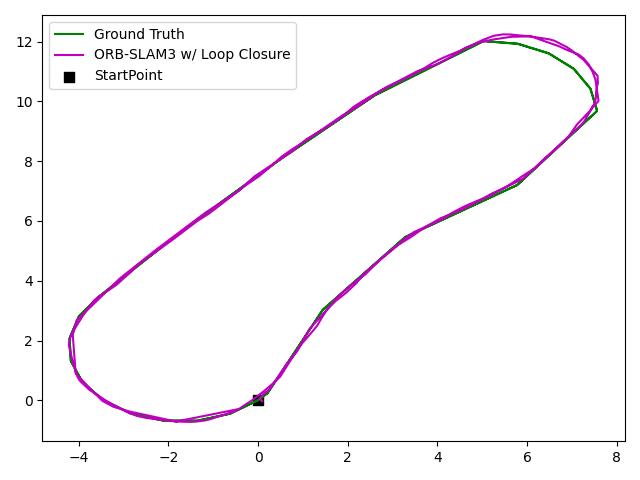}
    \hfill
    \includegraphics[width=0.48\linewidth]{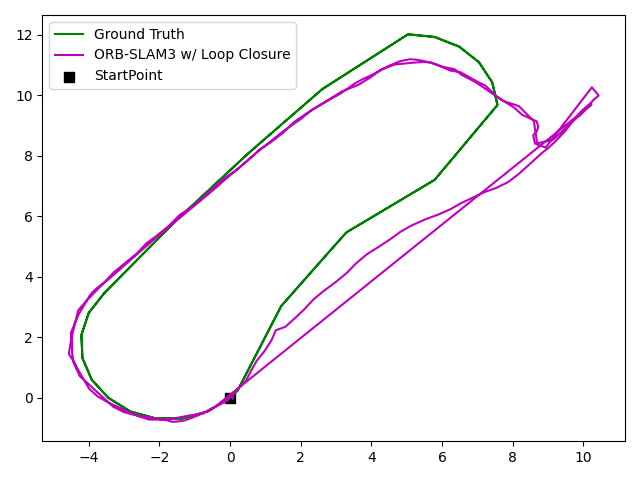}
    \caption{A comparison of ORB-SLAM3's performance on global shutter (left) and rolling shutter (right) sequences.}
    \label{fig:my_label}
\end{figure}

Our experiments all used global shutter images, as the odometry results are generally better with these. Figure~\ref{fig:my_label} compares ORB-SLAM3's estimated trajectories for global and rolling shutter images. The availability of simultaneously obtained images from these two types of cameras allows a quantitative evaluation of the relative value of one camera vs. the other. This significant difference highlights the need for explicit models for rolling shutter cameras~\cite{albl2019rolling} and the usefulness of BPOD to evaluate the effectiveness of such explicit models. 

\begin{figure}[t]
    \centering
    \includegraphics[width=0.48 \linewidth]{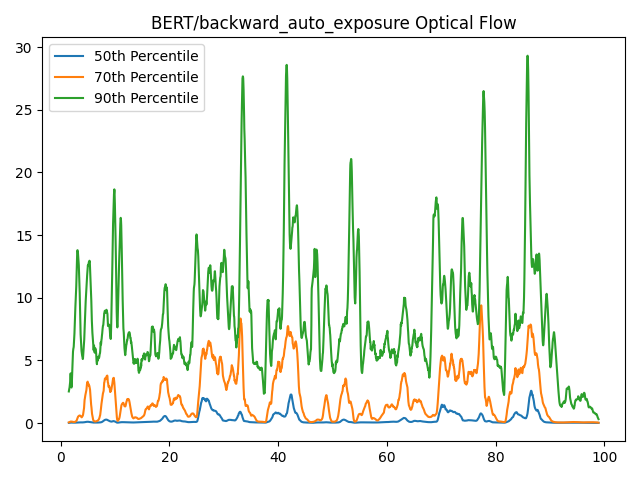}
    \hfill
    \includegraphics[width=0.48 \linewidth]{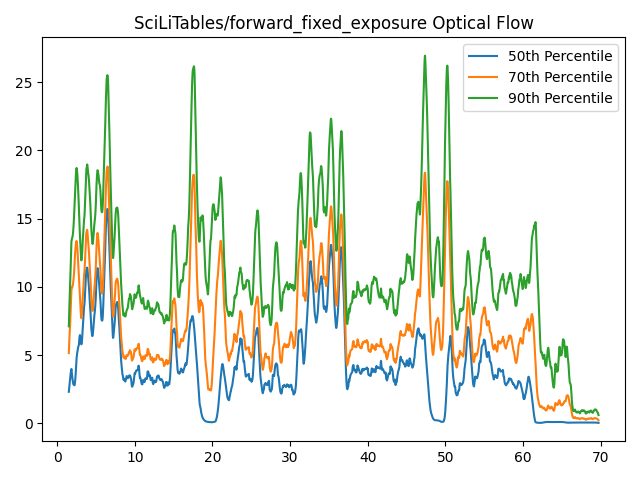}
    \includegraphics[width=0.48 \linewidth]{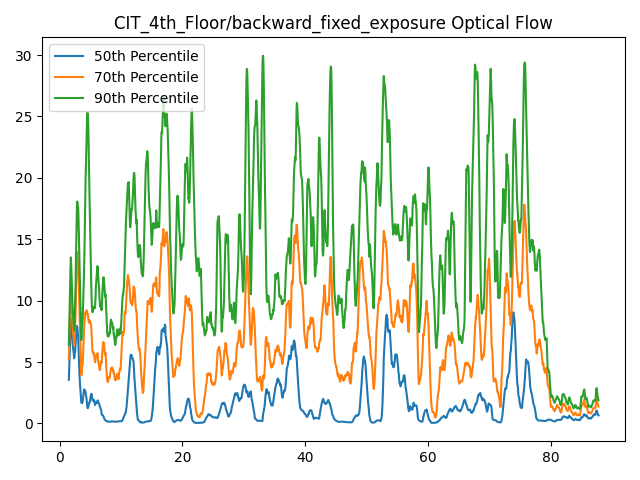}
    \hfill
    \includegraphics[width=0.48 \linewidth]{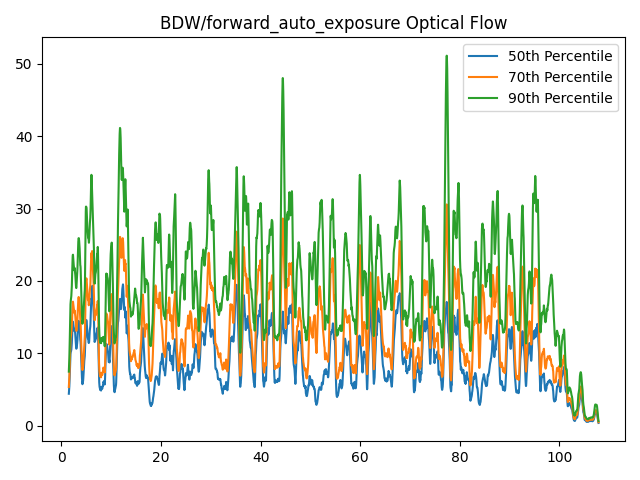}
    \caption{Plots of optical flow magnitude across our sequences show that BPOD is characterized by sudden, rapid movements that produce spikes in optical flow.}
    \label{fig:opticalFlow_single}
\end{figure}

Finally, the use of optical flow magnitude was proposed by~\cite{delmerico2019we} to visualize the level of difficulty across sequences in the same dataset. Figure~\ref{fig:opticalFlow_single} shows this measure over several sequences in BPOD, highlighting spikes of very large median (300 pixels/s) and 90th percentile (800 pixels/s) magnitudes.

\section{Conclusion}
We developed a novel head-mounted pedestrian dataset based on 12 locations that cover a diverse set of scenes and illumination conditions. The dataset contains synchronized stereo data from both rolling shutter and global shutter cameras, enabling a comparison of the two for future algorithm development, and is supplemented with IMU data. Tests on three representative feature-based, direct, and deep learning methods reveal that the dataset contains challenges not previously addressed by existing datasets. We hope that the BPOD dataset will facilitate the development of more reliable and accurate odometry methods to address the unique challenges of head-mounted pedestrian odometry.

{\small
\bibliographystyle{IEEEtran}
\bibliography{egbib}
}
\end{document}